\documentclass[11pt,a4paper]{article}
\usepackage[hyperref]{acl2021}
\usepackage{times}
\usepackage{latexsym}

\usepackage{graphicx}
\usepackage{xcolor}
\usepackage{multirow}
\usepackage{times}
\usepackage{url}
\usepackage{latexsym}
\usepackage{amssymb}
\usepackage{comment}

\newcommand{\bhline}{\noalign{\hrule height 1.5pt}}

\usepackage{microtype}

\aclfinalcopy 
\def\aclpaperid{3458} 

\title{Verb Sense Clustering using Contextualized Word Representations \\ 
for Semantic Frame Induction}

\author{
  Kosuke Yamada$^{1}$ \ \ \ \ \ \ \ \ \ Ryohei Sasano$^{1,2}$ \ \ \ \ \ \ \ \ \ Koichi Takeda$^{1}$  \\
  $^{1}$Graduate School of Informatics, Nagoya University, Japan \\
  $^{2}$RIKEN Center for Advanced Intelligence Project, Japan \\
  {\tt yamada.kosuke@c.mbox.nagoya-u.ac.jp}, \\
  {\tt \{sasano,takedasu\}@i.nagoya-u.ac.jp}
}

\date{}

\begin{document}
\maketitle
\begin{abstract}
Contextualized word representations have proven useful for various natural language processing tasks.
However, it remains unclear to what extent these representations can cover hand-coded semantic information such as semantic frames, which specify the semantic role of the arguments associated with a predicate.
In this paper, we focus on verbs that evoke different frames depending on the context, and we investigate how well contextualized word representations can recognize the difference of frames that the same verb evokes.
We also explore which types of representation are suitable for semantic frame induction.
In our experiments, we compare seven different contextualized word representations for two English frame-semantic resources, FrameNet and PropBank.
We demonstrate that several contextualized word representations, especially BERT and its variants, are considerably informative for semantic frame induction.
Furthermore, we examine the extent to which the contextualized representation of a verb can estimate the number of frames that the verb can evoke. 
\end{abstract}

\section{Introduction}
Contextualized word representations such as ELMo \cite{peters2018} and BERT \cite{devlin2019} are known to be effective in many natural language processing tasks such as question answering, natural language inference, and semantic textual similarity.
Contextualized word representations can generate different representations of the same word in different contexts and distinguish the polysemy of a word.
It has been reported that this property is effective in word sense disambiguation (WSD) \cite{hadiwinoto2019} and word sense induction (WSI) \cite{amrami2018}.
Therefore, it appears that contextualized word representations can also be leveraged to induce semantic frames from a large corpus automatically.

A semantic frame is defined on the basis of the semantic roles that a predicate can take as its arguments.
FrameNet\footnote{\url{https://framenet.icsi.berkeley.edu/}} \cite{baker1998} and PropBank\footnote{\url{https://propbank.github.io/}} \cite{palmer2005} are the two most well-known resources of semantic frames, both of which are manually compiled.
These resources are used not only for semantic parsing \cite{yang2017} but also for information extraction \cite{gangemi2016}, question answering \cite{shen2007}, and document summarization \cite{cheung2013}. 

These frame-semantic resources define the frame and semantic roles, and they provide example sentences in which they are annotated.
For example, the verb ``support'' in FrameNet is defined to evoke two frames: the \textsc{Supporting} frame and the \textsc{Evidence} frame.
Sentences (1) and (2) below are examples where these frames are annotated.
In Sentence (1), ``support'' means `supporting a person or a thing' and evokes the \textsc{Supporting} frame.
Its arguments are annotated with the semantic roles of \textsf{\small Supporter} and \textsf{\small Supported}.
In Sentence (2), ``support'' means `corroborating' and evokes the \textsc{Evidence} frame.
Its arguments are annotated with the semantic roles of  \textsf{\small Proposition} and  \textsf{\small Support}.
In both examples, the frame-evoking word is ``support,'' but its evoking frames are different.
\begin{itemize}
    \item[(1)] $[_{\rm \scriptsize \sf Supported}$ This study$]$
is \underline{supported} by $[_{\rm \scriptsize \sf Supporter}$ the fund$]$. (\textsc{Supporting})
    \item[(2)] $[_{\rm \scriptsize \sf Support}$ Our results$]$ \underline{support} $[_{\rm \scriptsize \sf Proposition}$ the hypothesis$]$. (\textsc{Evidence})
\end{itemize}

Since the manual development of such broad-coverage frame-semantic resources is labor-intensive and time-consuming, many researchers have attempted to induce semantic frames from large corpora automatically.
For example, \newcite{kawahara2014} extracted predicate-argument structures of each verb from large corpora and induced the frames that each verb evokes by clustering the extracted predicate-argument structures.
Several researchers have recently proposed frame induction methods that leverage word vector representations.
For example, \newcite{ustalov2018} collected subject-verb-object triples from a Web-scale corpus and induced the frames by clustering based on the concatenation of word vector representations of the triples.
However, since these approaches first collect the tuples of a verb and its arguments and then perform the clustering based on their word representations without taking their contexts into account, they may fail to disambiguate the word senses that require contextual clues.

\begin{figure}[!t]
    \centering
    \includegraphics[width=\linewidth]{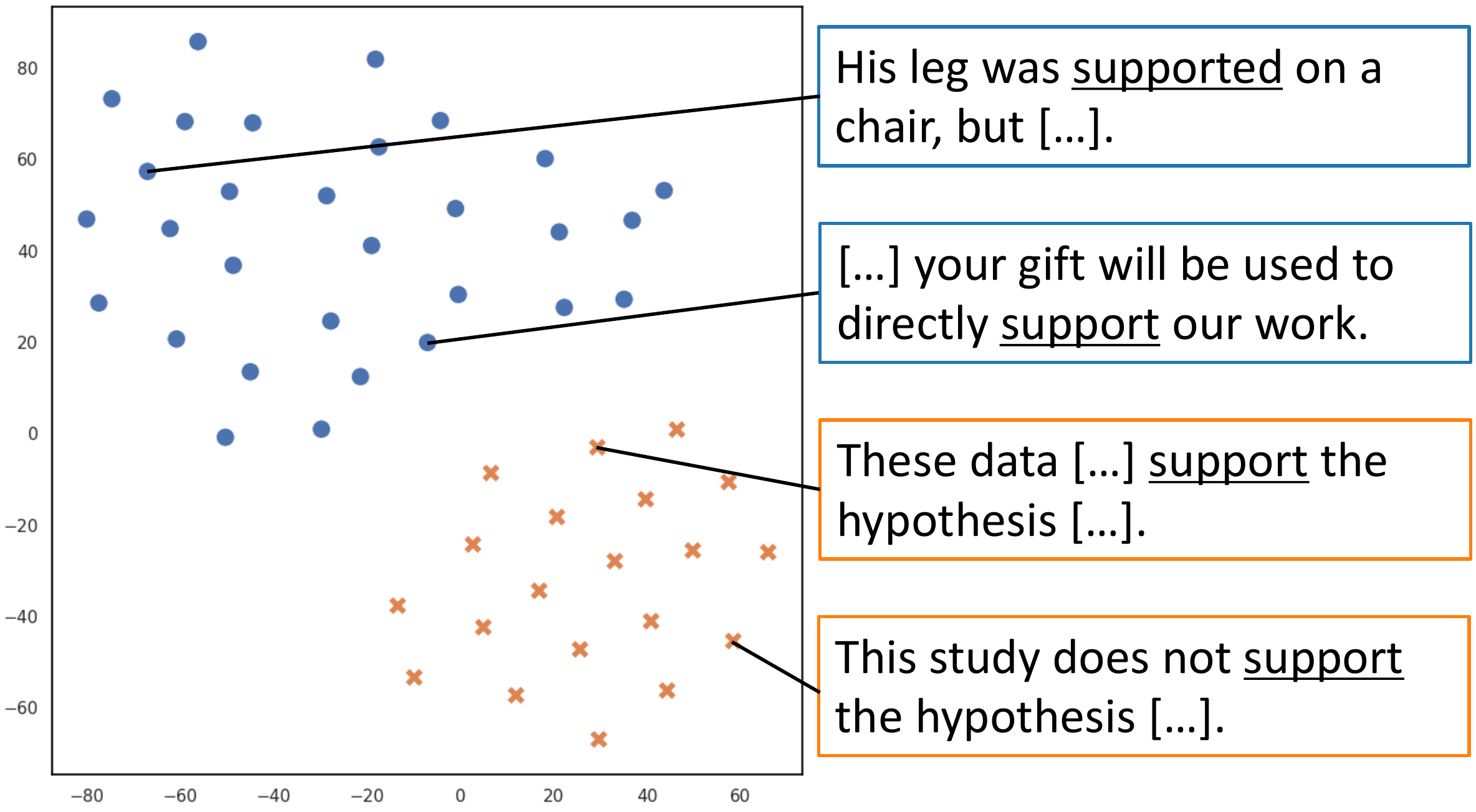}
    \caption{2D t-SNE projection of BERT representations of verb ``support'' in FrameNet. 
    {\Large\color[HTML]{1f77b4}$\bullet$} and {\color[HTML]{ff7f0e}$\times$} correspond to example sentences from \textsc{Supporting} and \textsc{Evidence} frames, respectively.
    }
    \label{fig:support_eg}
\end{figure}

Therefore, we seek a frame induction method that makes better use of contextual information by leveraging contextualized word representations.
Figure \ref{fig:support_eg} shows a 2D projection of contextualized representations of the verb ``support'' in different sentences.
We extracted example sentences of ``support'' from the frame-annotated sentences in FrameNet, acquired contextualized representations of the verbs by applying a pre-trained BERT, and then projected them into two dimensions by using t-distributed stochastic neighbor embedding (t-SNE) \cite{maaten2008}.
As shown in the figure, these BERT representations are distributed separately depending on the frame that ``support'' evokes in each example.

Our objective is to exploit this property of contextualized word representations for semantic frame induction.
As a first step, we investigate how well contextualized word representations can distinguish the frames that the same verb evokes and which type of representations are suitable for semantic frame induction.
We also need to estimate the number of frames that a verb evokes to build a frame-semantic resource automatically.
We clarify to what extent contextualized word representations of verbs can estimate the number of frames that verbs evoke, which are defined manually.
Our investigation of contextualized word representations will help construct high-quality frame-semantic resources not only for high-resource languages and general domains but also for low-resource languages and specific domains.

\section{Related Work}
\subsection{Contextualized Word Representations}
Contextualized word representations encode semantic and syntactic information by learning linguistic patterns and constraints from a large amount of text and provide significant improvements to the state of the art for a wide range of natural language processing tasks.
They are also widely applied as context-sensitive word representation extractors for summarization \cite{liu2019a}, neural machine translation \cite{zhu2019}, and so on.

Recently, several contextualized word representations have been proposed.
For example, ELMo \cite{peters2018} produces contextualized word representations by pre-training on a bidirectional language model task in 2-layer BiLSTMs. 
More recently, many Transformer-based \cite{vaswani2017} models have been proposed. 
BERT \cite{devlin2019} utilizes multilayer bidirectional Transformers and is pre-trained on two tasks: masked language modeling and next sentence prediction. 
RoBERTa \cite{liu2019b} redesigns the pre-training conditions for BERT, and ALBERT \cite{lan2019} shares each layer's parameters in BERT to reduce the number of parameters.
There are other models such as GPT-2 \cite{radford2019}, which is a unidirectional model that is trained to predict the next word in a sentence, and XLNet \cite{yang2019}, which is based on a permutation language model that learns a bidirectional context in an autoregressive manner.

\subsection{Semantic Frame Induction}
For semantic frame induction of a word in a context, it is a standard approach to extract predicate-argument structures and then perform the clustering of those structures. 
LDA-frames \cite{materna2012} is an approach that represents frames as tuples of subject and object and uses latent Dirichlet allocation (LDA) \cite{blei2003} to induce semantic frames. 
\newcite{kawahara2014} extracted predicate-arguments structures from a large Web corpus and then applied the Chinese restaurant process clustering-algorithm \cite{aldous1985} to group predicates with similar arguments. 
\newcite{ustalov2018} proposed the Triclustering, which produces subject-verb-object triples and then performs a graph-based clustering using the concatenations of their static word embeddings.
These methods take only the predicates and their arguments into account, and they do not sufficiently consider the context.

In some works, contextualized word representations are already used for semantic frame induction.
In a shared task at SemEval 2019 \cite{qasemizadeh2019}, some researchers worked on an unsupervised semantic frame induction task, and they reported that ELMo and BERT were useful for the task. 
\newcite{arefyev2019} first performed group average clustering by using contextualized word embeddings of target verbs from BERT.
Then, they performed clustering to split each cluster into two by using TF-IDF features with paraphrased words by using BERT.
\newcite{anwar2019} used a concatenated representation of a target verb and the average word embedding of all words in a sentence obtained by skip-gram \cite{mikolov2013} or ELMo.
They performed group average clustering based on Manhattan distance by using the embedding.
\newcite{ribeiro2019} performed graph clustering based on Chinese whispers \cite{biemann2006} by using contextualized representations of frame-evoking verbs from ELMo or BERT.

The shared task dataset contains many example sentences in which different verbs evoke the same frame, and thus the dataset is suitable for evaluating semantic frame induction over verbs.
However, there are few example sentences of verbs that evoke different frames in the dataset, and it is not ideal for analyzing the difference of frames that each verb evokes. 
Some researchers assumed that many verbs evoke only one frame, and they did not analyze the difference of frames that each verb evokes. 

Also, there is a study that works on semantic frame induction by using contextualized word representations in semi-supervised learning.
\newcite{yong2020} used ELMo or BERT and mapped high-dimensional representations of verbs to a low-dimensional latent space for better frame prediction.
Their study aims to extend FrameNet.
On the other hand, our goal is to build frame-semantic resources automatically in an unsupervised fashion.

\subsection{Word Sense Disambiguation with Contextualized Word Representation}
The task in this paper is to distinguish the difference of frames that the same verb evokes, and as such, can be regarded as a type of word sense disambiguation (WSD) task. 
For the WSD task, contextualized word representations have been reported to be useful.
For example, \newcite{peters2018} performed the task by nearest neighbor matching with ELMo representations, and \newcite{hadiwinoto2019} used pre-trained BERT contextualized representations as features for WSD. 
While WSD aims to distinguish between the meanings of words on the same surface, the semantic frame induction we focus on aims to distinguish between intuitive concepts such as situations, objects, and events that words evoke.

\section{Methodology}
We investigate to what extent contextualized word representations recognize the difference of frames that the same verb evokes. 
Specifically, we focus on verbs that evoke more than one frame in frame-semantic resources and acquire contextualized word representations of them.
We then apply clustering and evaluate how well the generated clusters and human-annotated frames match.

\subsection{Frame-semantic Resources}
We use FrameNet and PropBank in English as frame-semantic resources.
Since our goal is to establish a semantic frame induction method that is not in a particular style, we use two well-known frame knowledge resources for our investigation: Berkeley FrameNet data release 1.7\footnote{\url{https://framenet.icsi.berkeley.edu/fndrupal/framenet_data}} and PropBank-annotated data from OntoNotes v5.0.\footnote{\url{https://catalog.ldc.upenn.edu/LDC2013T19}}

FrameNet is developed within the framework of the theory of frame semantics proposed by \newcite{fillmore2006}.
Each frame is shared by multiple frame-evoking words (lexical units), and hierarchical relations such as ``Inheritance'' or ``Using'' are defined between closely related frames.
FrameNet has 1,222 frames, 13,572 lexical units, and 200,751 annotated sentences.
The corpus consists of the British National Corpus and U.S. newswire texts.

PropBank is developed as a corpus with semantic roles that can be used as training data in supervised learning.
PropBank frames are defined for each verb as a frameset containing semantic role labels.
There are two types of labels; one is ARG0-5. 
It indicates a necessary role and has a different meaning in each frameset. 
The other is an argument modifier (AM) label, which indicates an additional role common to all framesets (e.g., AM-TMP for time).
For example, the frameset \textsc{support.01} (lend aid, credence to) of ``support'' is defined with ARG0 as `helper' and ARG1 as `person, thing or project being supported.'
Sentence (3) is an example in which this frameset is annotated.
\begin{itemize}
    \item[(3)] $[_{\rm \scriptsize ARG1}$ Students$]$
were \underline{supported}$_{\rm \scriptsize .01}$ by $[_{\rm \scriptsize ARG0}$ the scholarship$]$ $[_{\rm \scriptsize AM-TMP}$ for four years$]$.
\end{itemize}

\noindent
Unlike FrameNet, hierarchical relations are not defined between framesets; that is, each frameset is independent.
PropBank has 5,607 framesets, 4,221 verbs, and 111,178 annotated sentences.
The corpus consists of newswires, magazine articles, broadcast news, broadcast conversations, web data, conversational speech data, and pivot text. 

\subsection{Procedure}
In our investigation, we follow the procedures below for each target verb that evokes more than one frame in the frame-semantic resources.
\begin{enumerate}
\item 
Acquire contextualized word representations of the target verbs in the set of frame-annotated example sentences in the frame-semantic resources.\footnote{
Tokenization is performed in the same way as used in the pre-training.
If tokenization splits the target verb token into more than one sub-token, we use the contextualized word representations of the first sub-token.}
\item 
Apply clustering to their contextualized word representations by using a Gaussian mixture model. 
At this time, the number of clusters given to the model is equal to the number of frames in our dataset.
\item 
Find a mapping between the generated clusters and the human-annotated frames that maximize the overall number of matches.
We use the match rate as the evaluation metrics.
\end{enumerate}  

\section{Experiment on Frame Distinction}
\label{section_ex}
\subsection{Dataset}
\label{sec:dataset}
We first determined the target verbs, and we then extracted example sentences of the target verbs from both FrameNet and PropBank. 
As target verbs, we used verbs that evoke two or more frames with at least 20 annotated sentences. 
For example, in FrameNet, the verb ``support'' is a target verb because there are 30 sentences in the \textsc{Supporting} frame and 20 sentences in the \textsc{Evidence} frame.
In contrast, the verb ``attend'' is not a target verb. 
This is because although the verb ``attend'' evokes three frames, (\textsc{Attention}, \textsc{Perception\_active}, and \textsc{Attending}), there are 7 sentences in the \textsc{Attention} frame, 4 sentences in the \textsc{Perception\_active} frame, and 24 sentences in the \textsc{Attending} frame and only the \textsc{Attending} frame includes 20 or more sentences.

\begin{table*}[!t]
    \centering
    \small
    \begin{tabular}{l|cccc@{\ \ \ \ \ \ \ \ }r@{}c@{}l} \bhline
    Model & Corpus size & No. of parameters & No. of dimensions & No. of layers & \multicolumn{3}{@{\hspace*{-3mm}}c@{\ }}{Used layer} \\ \hline
    ELMo                & \ \ 11GB & \ \ 94M & \ \ 512 & \ \ 2 &  1 & $|$ &  1 \\
    BERT$_{\rm{BASE}}$  & \ \ 16GB &    110M & \ \ 768 &    12 &  9 & $|$ &  7 \\
    BERT$_{\rm{LARGE}}$ & \ \ 16GB &    340M &    1024 &    24 & 21 & $|$ & 15 \\
    RoBERTa             &    160GB &    125M & \ \ 768 &    12 & 10 & $|$ &  6 \\ 
    ALBERT              & \ \ 16GB & \ \ 11M & \ \ 768 &    12 &  9 & $|$ &  8 \\ 
    GPT-2               & \ \ 40GB &    117M & \ \ 768 &    12 &  8 & $|$ &  9 \\ 
    XLNet               &    158GB &    110M & \ \ 768 &    12 &  5 & $|$ &  5 \\
    \bhline
    \end{tabular}
    \caption{Details of contextualized word representations. 
    ``Used layer'' means hidden layer of model used to obtain the representations in FrameNet (left) and PropBank (right).}
    \label{tb:model}
\end{table*}

For each verb, we considered frames that include at least 20 annotated sentences. 
In addition, if the target verb evokes more than 10 frames with 20 or more annotated sentences, we used the top 10 frames on the basis of the number of annotated sentences.
We used a maximum of 100 annotated sentences for each frame. 
As a result, we have obtained 178 target verbs for FrameNet and 164 for PropBank. 
The average counts of frames per verb were 2.21 for FrameNet and 2.73 for PropBank, and the average counts of annotated sentences per frame were 41.68 for FrameNet and 70.34 for PropBank.
In this paper, we used 120 verbs as the test set for the final evaluation and the remaining verbs as the development set for tuning the parameters for both FrameNet and PropBank.

\subsection{Settings}
We compared ELMo, BERT$_{\rm{BASE}}$, BERT$_{\rm{LARGE}}$, RoBERTa, ALBERT, GPT-2, and XLNet as contextualized word representations in order to explore the representation most suitable for semantic frame induction. 
We used publicly available pre-trained models. 
ELMo is the `Original' model in AllenNLP,\footnote{\url{https://allennlp.org/elmo}} and the other transformer-based models are pre-trained models\footnote{
These models are specified by `bert-base-uncased,' `bert-large-uncased,' `roberta-base,' `albert-base-v2,' `gpt2,' and `xlnet-base-cased' in Hugging Face.
} in Hugging Face.\footnote{\url{https://huggingface.co/transformers/}}
For each model, we obtained contextualized word representations from the hidden layer that achieved the highest scores in the development sets for FrameNet and PropBank, respectively.
Table \ref{tb:model} lists the size of the corpus used to pre-train models and the number of parameters, dimensions, layers, and hidden layers of models used to obtain the representations for FrameNet and PropBank, respectively.

We used the Gaussian mixture model implementation provided by scikit-learn.\footnote{\url{ https://scikit-learn.org}}
We adopted ``spherical'' as the covariance type, that is, the covariance matrix was a diagonal covariance with equal elements along the diagonal.
We used five trials of clustering with different random seeds and adopted the result of the highest likelihood trial. 

\begin{table}[!t]
    \centering
    \small
    \begin{tabular}{l|cc} \bhline
    Model               & FrameNet       & PropBank \\ \hline
    All-in-one-cluster & 0.578 & 0.548\\ 
    ELMo & 0.631 & 0.607 \\
    BERT$_{\rm{BASE}}$ & 0.750 & 0.765 \\
    BERT$_{\rm{LARGE}}$ & \textbf{0.769} & 0.790 \\
    RoBERTa & 0.767 & \textbf{0.796} \\ 
    ALBERT & 0.705 & 0.712 \\
    GPT-2 & 0.666 & 0.650 \\
    XLNet & 0.729 & 0.758 \\ \bhline
    \end{tabular}
    \caption{Macro-average match rate of each verb for each of models and datasets.}
    \label{TB:SCORE}
\end{table}

\subsection{Results}
\label{SEC:EX}
Table \ref{TB:SCORE} lists the macro-average match rate of each verb for each of the models and datasets.
All-in-one-cluster means the average score when all the example sentences were in one cluster for each verb.
That is, the score is the average of the percentages of examples that were annotated with the most frequently used frame for a verb.
For example, the score of the verb ``support'' in FrameNet was 0.6 (30/50) since the numbers of example sentences from the \textsc{Supporting} frame and the \textsc{Evidence} frame were 30 and 20, respectively.

As shown in Table \ref{TB:SCORE}, BERT$_{\rm{LARGE}}$ and RoBERTa achieved the highest scores for FrameNet and PropBank, respectively. 
We confirmed that they recognized the differences of frames that the same verbs evoke.
BERT$_{\rm{BASE}}$, XLNet, and ALBERT also achieved high scores.
These results indicate that BERT, RoBERTa, XLNet, and ALBERT are useful for semantic frame induction.
In contrast, the scores obtained for ELMo and GPT-2 were relatively low and almost the same as for the All-in-one-cluster.
It indicates that the degree of the difference of frames captured by the contextualized word representations varied greatly.

The reasons for these results are described below.
The high scoring BERT, RoBERTa, XLNet, and ALBERT are deep bidirectional language models based on Transformer. 
In contrast, GPT-2 is a unidirectional language model based on Transformer.
Also, ELMo is a relatively sparse bidirectional language model that consists of only two unidirectional contexts concatenated together.
Therefore, the scores of GPT-2 and ELMo were lower than those of the deep bidirectional language models.

\begin{figure}[!t]
    \centering
    \includegraphics[width=\linewidth]{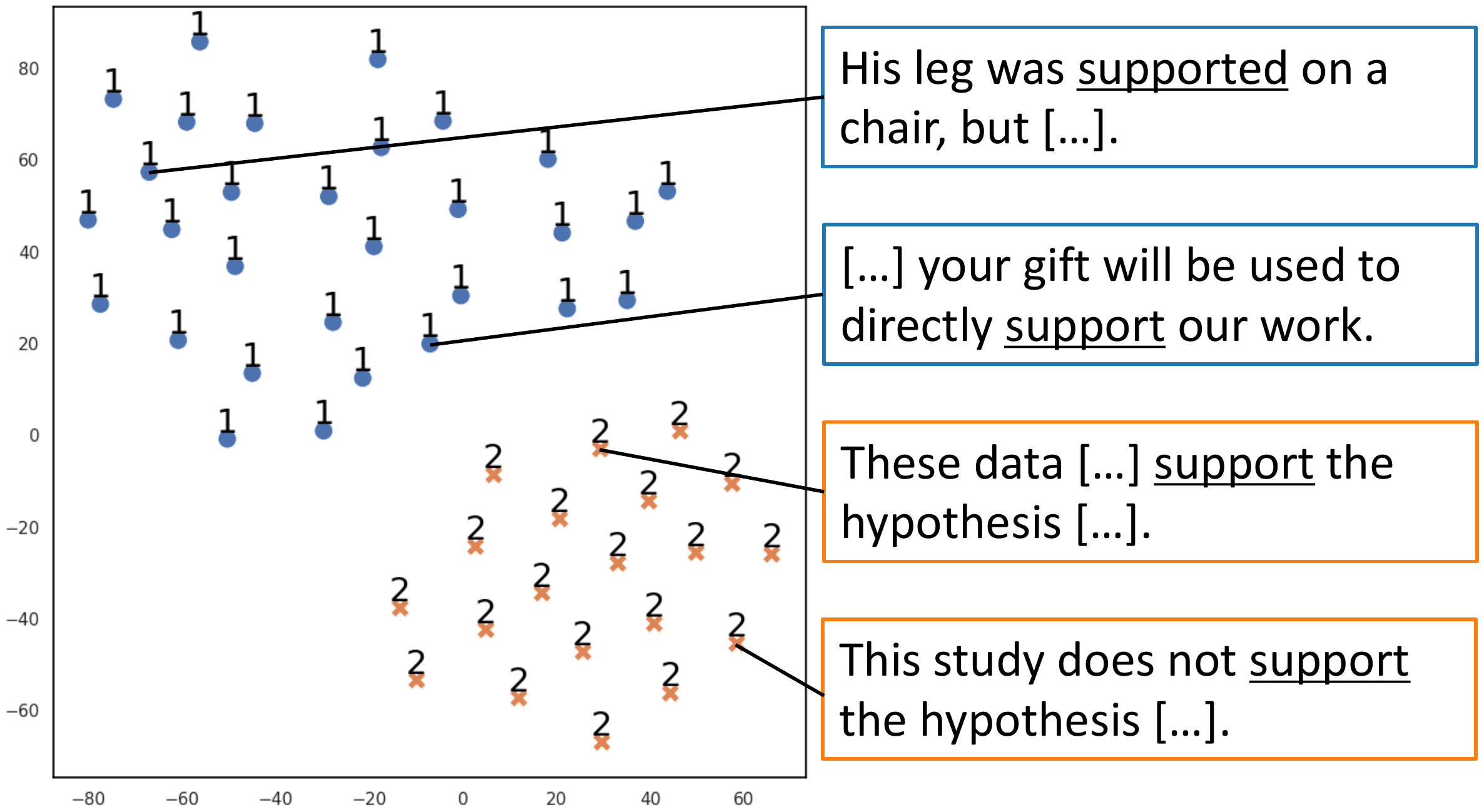}
    \caption{
    2D t-SNE projection of BERT$_{\rm{LARGE}}$ vectors for verb ``support'' in FrameNet. 
    {\Large\color[HTML]{1f77b4}$\bullet$} and {\color[HTML]{ff7f0e}$\times$} correspond to example sentences from \textsc{Supporting} and \textsc{Evidence} frames, respectively.}
    \label{fig:support}
\end{figure}

We show several examples below. 
In these figures, the number given to each point represents the clustering result; that is, the points with the same number belong to the same cluster. 
Note that the value of the number has no meaning. 
Figure \ref{fig:support} shows a 2D t-SNE projection of BERT$_{\rm{LARGE}}$ vectors for ``support” in FrameNet.
We can see that the example sentences from the \textsc{Supporting} frame and the \textsc{Evidence} frame form a cluster, respectively.

\begin{figure}[!t]
    \centering
    \includegraphics[width=\linewidth]{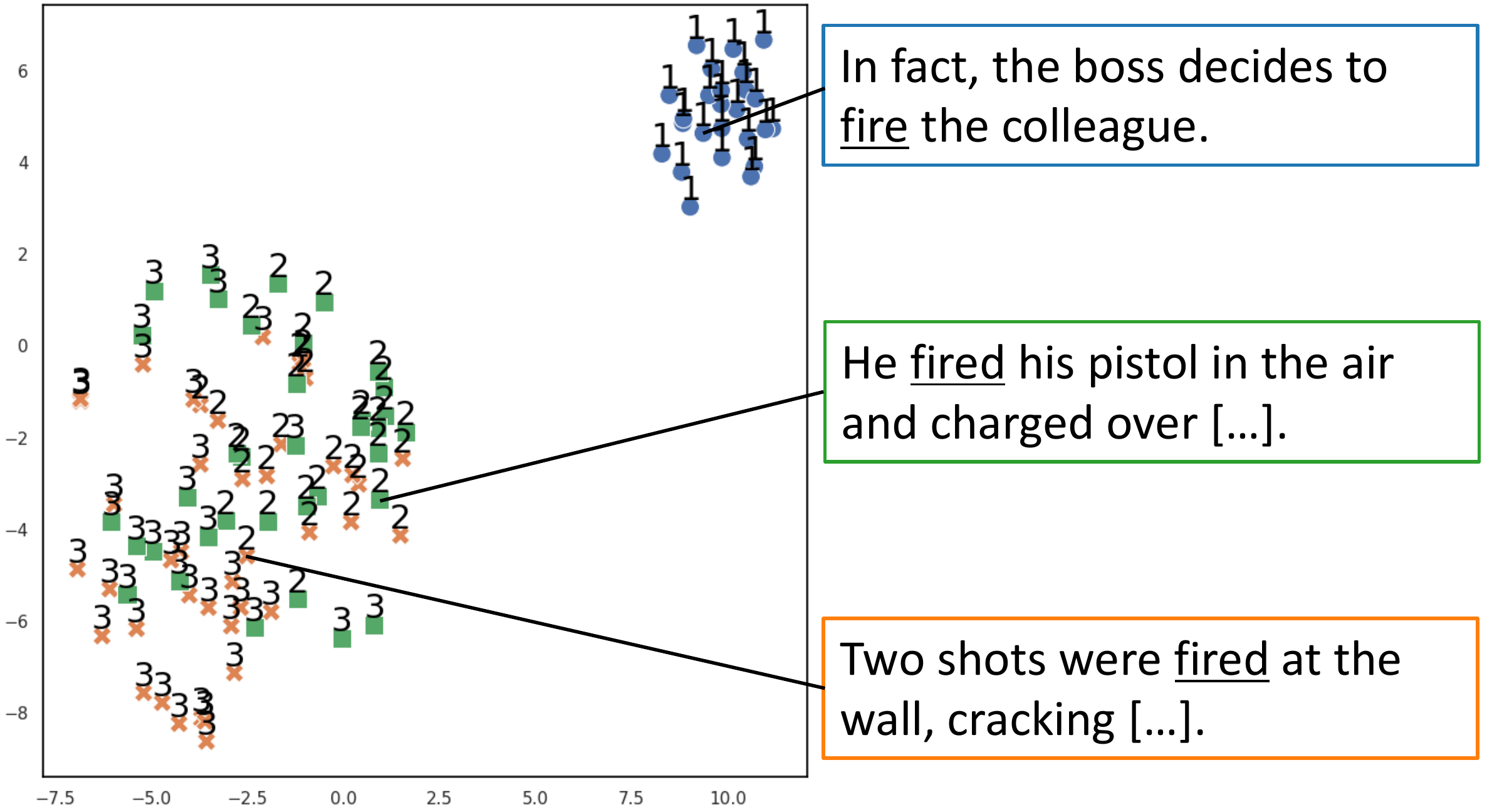}
    \caption{
    2D t-SNE projection of BERT$_{\rm{LARGE}}$ vectors for verb ``fire'' in FrameNet.
    {\Large \color[HTML]{1f77b4}$\bullet$}, {\color[HTML]{ff7f0e}$\times$}, and {\small \color[HTML]{2ca02c}$\blacksquare$} correspond to example sentences from \textsc{Firing}, \textsc{Shoot\_projectiles}, and \textsc{Use\_firearm} frames, respectively.}
    \label{fig:fire}
\end{figure}

Figure \ref{fig:fire} shows a 2D t-SNE projection of BERT$_{\rm{LARGE}}$ vectors for ``fire'' in FrameNet.
We can see that example sentences from the \textsc{Firing} frame form a single cluster, whereas the difference between the \textsc{Shoot\_projectiles} frame and the \textsc{Use\_firearm} frame is not captured.
The \textsc{Firing} frame means `ending an employment relationship' while the \textsc{Shoot\_projectiles} and the \textsc{Use\_firearm} frames mean `shooting a bullet' and `shooting a gun', respectively. 
The \textsc{Firing} frame is very different from the other two. 
On the other hand, the ``Using'' relation is annotated between the \textsc{Shoot\_projectiles} frame and \textsc{Use\_firearm} frame, which indicates that there is a strong connection between the two frames. 
We conduct an additional analysis on frames that have hierarchical relations in Section \ref{sec:relation}.

\begin{figure}[!t]
    \centering
    \includegraphics[width=\linewidth]{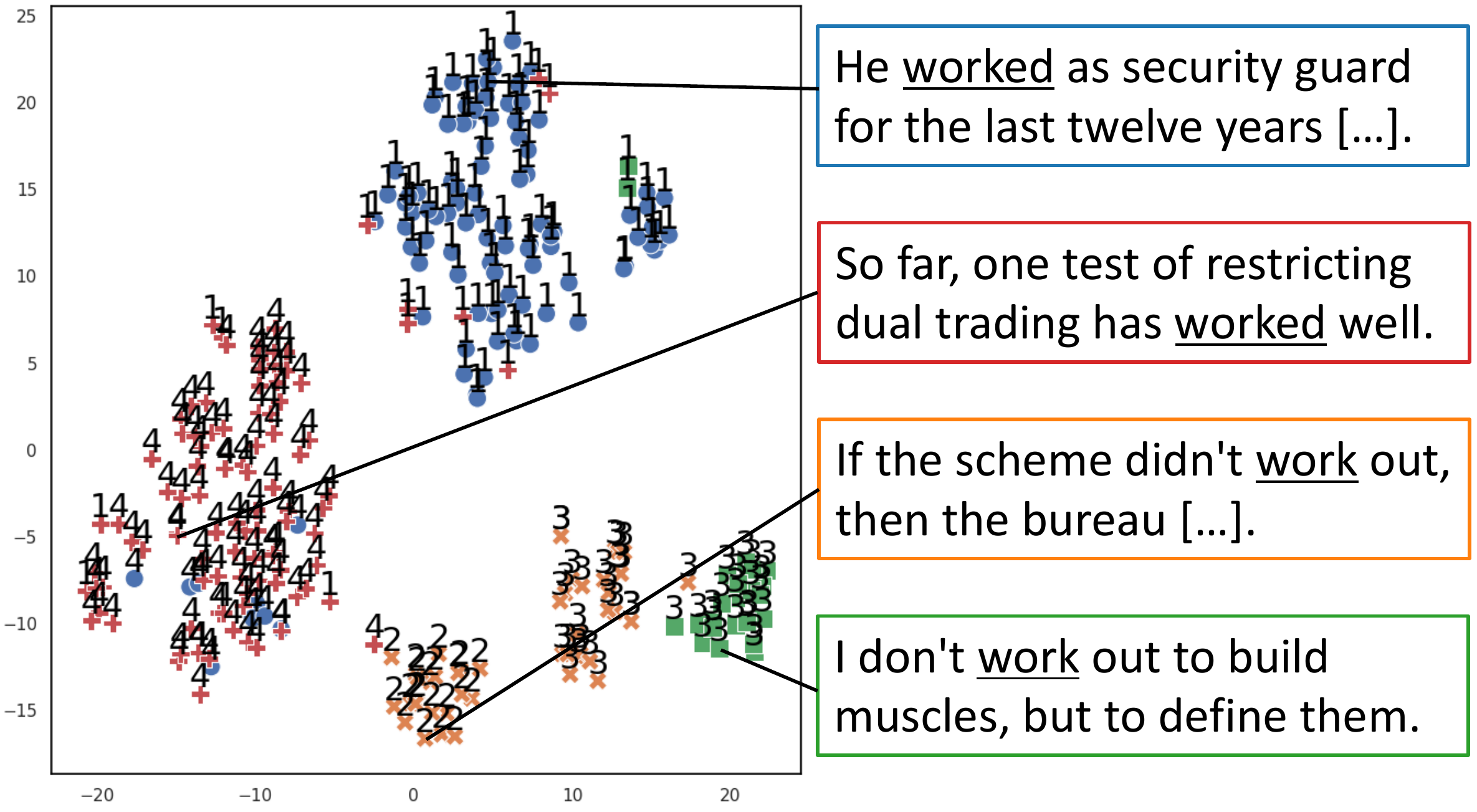}
    \caption{
    2D t-SNE projection of BERT$_{\rm{LARGE}}$ vectors for verb ``work'' in PropBank.
    {\Large \color[HTML]{1f77b4}$\bullet$}, {\color[HTML]{ff7f0e}$\times$}, {\small \color[HTML]{2ca02c}$\blacksquare$}, and {\color[HTML]{d62728}$+$} correspond to example sentences from \textsc{work.01} (work), \textsc{work.02} (arrange), \textsc{work.03} (exercise), and  \textsc{work.09} (function, operate) framesets, respectively.}
    \label{fig:work}
\end{figure}

Figure \ref{fig:work} shows a 2D t-SNE projection of BERT$_{\rm{LARGE}}$ vectors for ``work'' in PropBank. 
The verb ``work'' has four types of framesets: \textsc{work.01} (work), \textsc{work.02} (arrange), \textsc{work.03} (exercise), and \textsc{work.09} (function, operate). 
We confirmed that BERT$_{\rm{LARGE}}$ roughly captured the difference of frames, even for verbs that can have many framesets.
In the examples where \textsc{work.02} and \textsc{work.03} were annotated, the verb ``work'' appears in the form of ``work out,'' and it may have been a bit challenging to capture the difference of these framesets.
This is because verbs that appear as part of phrasal verbs have relatively similar contextualized word embeddings since the same word appears near the verb.

\begin{figure}[!t]
    \centering
    \includegraphics[width=\linewidth]{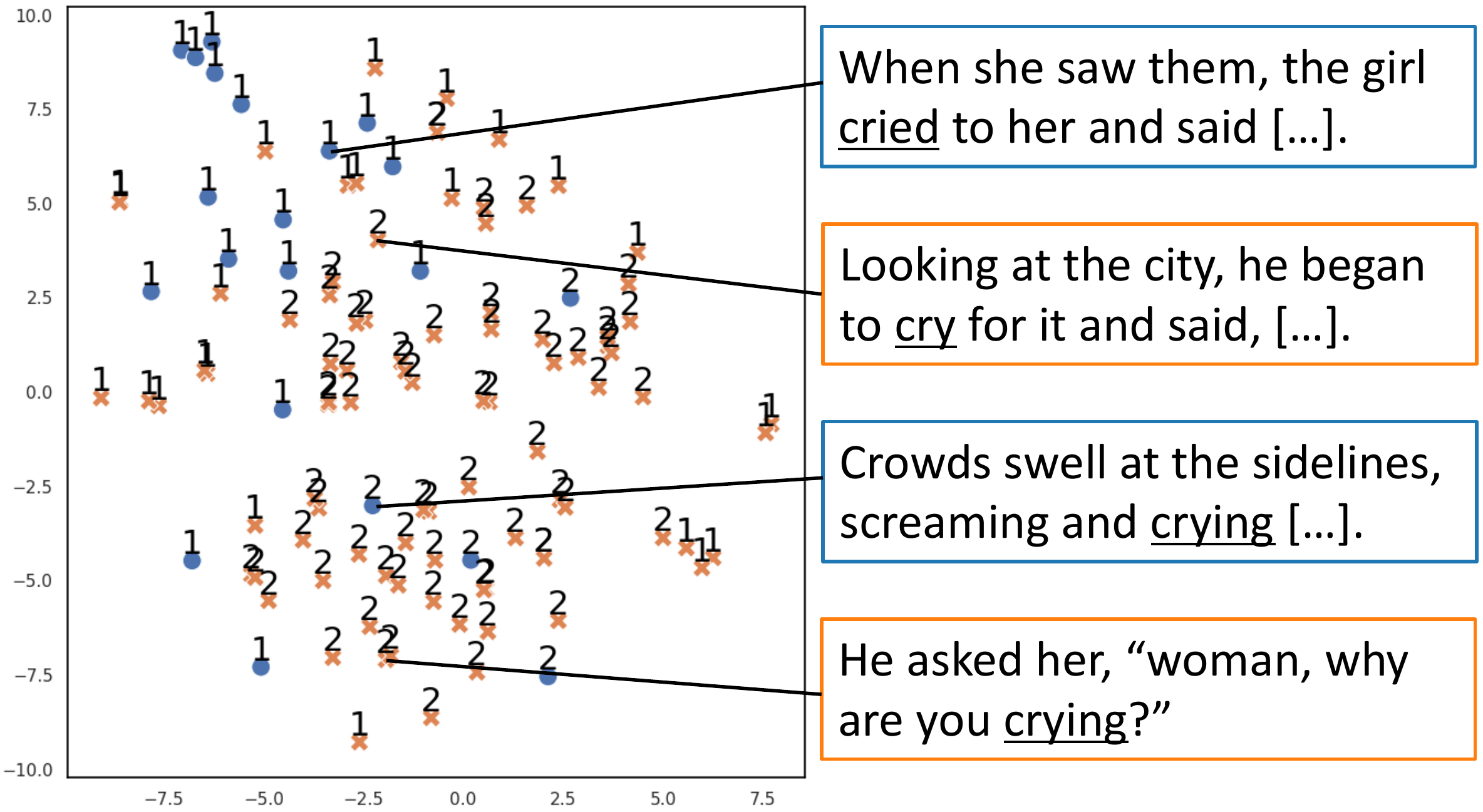}
    \caption{
    2D t-SNE projection of BERT$_{\rm{LARGE}}$ vectors for verb ``cry'' in PropBank.
    {\Large \color[HTML]{1f77b4}$\bullet$} and {\color[HTML]{ff7f0e}$\times$} correspond to example sentences from \textsc{cry.01} (speak loudly, yell, demand, possibly while weeping) and \textsc{cry.02} (cry, weep) framesets, respectively.}
    \label{fig:cry}
\end{figure}

Figure \ref{fig:cry} shows a 2D t-SNE projection of BERT$_{\rm{LARGE}}$ vectors for ``cry'' in PropBank. 
The verb ``cry'' has two types of framesets: \textsc{cry.01} (speak loudly, yell, demand, possibly while weeping) and \textsc{cry.02} (cry, weep). 
Like the verb ``fire'' in FrameNet, the resulting clusters could not be appropriately formed because the framesets of the verb ``cry'' are both related to `weep' and are thus very similar.

\subsection{Effect of Hierarchical Relations on Evaluation}
\label{sec:relation}
The frames with hierarchical relations defined in FrameNet appear in similar contexts.
As is clear from the examples of ``fire,'' it is not easy to distinguish these frames, even using contextual word representations. 
Moreover, it is unclear whether these frames should be defined as separate frames if semantic frame resources are to be automatically constructed in the future. 
Specifically, the importance of distinguishing between the \textsc{Shoot\_projectiles} frame and the \textsc{Use\_firearm} frame could be less important than distinguishing between the \textsc{Shoot\_projectiles} frame and the \textsc{Firing} frame.

To investigate the practical usefulness, we attempted to evaluate the accuracy of the distinction between frames with hierarchical relations and frames without relations, separately.
We first extracted verbs that had exactly two types of frames from FrameNet as a result of the procedure described in Section \ref{sec:dataset}.
We then divided the extracted verbs into two groups according to whether there is a hierarchical relation between the two frames or not and calculated the average match rate for each group.\footnote{
Among the 120 verbs that were used as the test set in the experiment, the number of verbs that evoked only two frames was 96.
The number of verbs in the group without the relations was 62,
while the number of verbs in the group with the relations was 34.}
By limiting our focus to verbs with two types of frames, we can ignore the tendency of the match rate to decrease as the number of frames increases.
We assume that if a certain contextualized word representation appropriately captures the difference of frames, it should be able to distinguish the difference of frames with a high match rate.

\begin{table}[!t]
    \centering
    \small
    \begin{tabular}{l|cc|r} \bhline
    Model               & Gr w/o rel & Gr w/ rel  & \multicolumn{1}{c}{Diff} \\ \hline
    All-in-one-cluster & 0.613 & 0.604 & 0.009 \\ 
    ELMo & 0.680 & 0.644 & 0.036 \\
    BERT$_{\rm{BASE}}$ & 0.805 & 0.719 & 0.086 \\ 
    BERT$_{\rm{LARGE}}$ & 0.826 & 0.713 & 0.113  \\
    RoBERTa & 0.836 & 0.707 & 0.129 \\ 
    ALBERT & 0.758 & 0.670 & 0.088 \\ 
    GPT-2 & 0.724 & 0.655 & 0.069 \\
    XLNet & 0.783 & 0.700 & 0.083 \\ \bhline
    \end{tabular}
    \caption{
    Average match rate by groups without hierarchical relations (Gr w/o rel) and groups with hierarchical relations (Gr w/ rel). 
    ``Diff'' represents difference between score of Gr w/o rel and score of Gr w/ rel.}
    \label{TAB:Relation}
\end{table}

Table \ref{TAB:Relation} lists the results of the average match rate.
In the models of BERT$_{\rm{LARGE}}$, RoBERTa, BERT$_{\rm{BASE}}$, XLNet, and ALBERT, which obtained relatively high scores in the results shown in Table \ref{TB:SCORE}, we can see that the group without relations got higher scores than the group with relations. 
It is arguably concluded this result indicates that these models accurately captured the essential difference of frames.

\section{Estimation of Number of Frames}
In the experiments in Section \ref{section_ex}, we gave the number of frames in our dataset to the Gaussian mixture model.
However, it is necessary to estimate the number of frames that each verb evokes for semantic frame induction.
Therefore, we investigated how well we can estimate the number of frames on the basis of information criteria by using contextualized word representations.
Specifically, we adopted a Bayesian information criterion (BIC) \cite{schwarz1978}, which is used for determining the number of clusters, and an adjusted-BIC, in which the BIC is adjusted so that the estimated number of clusters is close to the number of human-annotated frames.

\subsection{Information Criterion}
The Bayesian information criterion (BIC) is one of the most widely used criterion for model selection. The BIC is defined as
\begin{equation}
\mathrm{BIC} = -2\mathrm{ln}(L) + k \cdot \mathrm{ln}(n_s),
\label{eq::bic}
\end{equation}
\noindent
where $L$ is the likelihood of the model, $n_s$ is the number of samples, and $k$ is the number of model parameters. 
The parameters of the Gaussian mixture model consist of the mean, covariance, and mixture weights. 
When the numbers of clusters and dimensions are represented by $n_c$ and $d$, respectively, the number of parameters required to represent the mean is $d \times n_c$. 
Since we adopted spherical as the covariance type, where each component has its own single variance, the number of parameters required to represent the covariance is $n_c$. 
Since the mixture weights for each component are probabilities that sum to 1, the number of parameters required to represent the mixture weight is $n_c-1$. 
Thus, the total number of model parameters is $k=(d+2) \times n_c-1$. 
When the BIC is used to determine the number of clusters, the number that minimizes the BIC is selected.

The first term on the right-hand side in Equation \ref{eq::bic} decreases as the number of clusters increases because the likelihood of an optimized model generally increases as the number of parameters increases. 
The second term on the right-hand side is regarded as a penalty term that inhibits the increase in the number of clusters.
The granularity of frames decided by human intuition may not be optimal in terms of the information criterion. 
Therefore, we introduce an adjusted-BIC in which the penalty term of the BIC is adjusted so that the granularity of the frames is close to human intuition.
The equation of the adjusted-BIC (a-BIC) is defined as
\begin{equation}
\mathrm{a\mathchar`-BIC} = -2\mathrm{ln}(L) + c \cdot k \cdot \mathrm{ln}(n_s),
\end{equation}
\noindent
where $c$ is a constant that adjusts the penalty, which is decided by using the development set.\footnote{
The constant $c$ is increased from 1 in increments of 0.1, and we decide the value when the total number of frames and the total estimated number of clusters are as close as possible in the development set.}

\subsection{Results}
In the experiments in Section \ref{section_ex}, we used only the verbs that evoke more than one type of frame.
However, it is also essential for verbs that evoke only one type of frame to recognize that.
Therefore, we added verbs that evoke only one frame. 
The number of verbs added was the same as the number of verbs used in the experiment in Section \ref{section_ex}. 
We also used a maximum of 100 annotated sentences from each frame. 
As a result, we used 116 verbs for parameter tuning as the development set and 240 verbs for evaluation as the test set for FrameNet, and we used 88 verbs for parameter tuning as the development set and 240 verbs for evaluation as the test set for PropBank.

We used BERT$_{\rm{LARGE}}$ as contextualization word representations. 
We evaluated the automatic estimation of the number of frames by using Spearman's rank correlation coefficient ($\rho$), accuracy, and root mean square error (RMSE) for the estimated number of clusters and the number of frames in our dataset.
Table \ref{TB:CNUM} lists the estimation results of the number of frames.
For both FrameNet and PropBank, using the adjusted-BIC as the information criterion resulted in better scores than using the BIC. 
When using the adjusted-BIC, Spearman's rank correlation coefficients were 0.177 and 0.631 for FrameNet and PropBank, respectively. 
The accuracy scores were over 0.5, which means that we could correctly predict the number of frames for more than half of the verbs.
The accuracy for FrameNet is lower than the accuracy for PropBank.
Accurate prediction of the number of frames for FrameNet will need to consider semantic coherence across different verbs, since frames in FrameNet are not defined independently for each verb.

\begin{table}[t!]
    \centering
    \small
    \begin{tabular}{ll|ccc} \bhline
    & & \multicolumn{1}{c}{$\rho$} & \multicolumn{1}{c}{Accuracy} & \multicolumn{1}{c}{RMSE} \\ \hline
    \multirow{2}{*}{FrameNet} & BIC & 0.165 & 0.058 & 3.131 \\
                        {} & a-BIC & 0.177 & 0.517 & 1.195 \\ \hline
    \multirow{2}{*}{PropBank} & BIC & 0.066 & 0.004 & 7.429 \\
                        {} & a-BIC & 0.631 & 0.608 & 1.008 \\ \bhline
    \end{tabular}
    \caption{Evaluation on estimating the number of frames using BIC and a-BIC for FrameNet and PropBank.}
    \label{TB:CNUM}
\end{table}

Table \ref{TB:CM} shows the confusion matrices between the number of human-annotated frames and the estimated number of frames using the BIC and the adjusted-BIC for FrameNet and PropBank. 
We can see that the BIC tended to overestimate the number of frames. 
The constant c was tuned at 3.1 for FrameNet and 3.4 for PropBank.

\section{Conclusion and Future Work}
We investigated to what extent contextualized word representations can recognize the difference of frames that the same verb evokes.
Specifically, we focused on verbs that evoke multiple frames and performed clustering based on contextualized word representations of target verbs.
We calculated the match rate between the generated clusters and the human-annotated frames and compared seven contextualized word representations: ELMo, BERT$_{\rm{BASE}}$, BERT$_{\rm{LARGE}}$, RoBERTa, ALBERT, GPT-2, and XLNet.
We found that BERT, RoBERTa, XLNet, and ALBERT achieved high performance in distinguishing the difference of frames that the same verb evokes.
We also found that we can estimate the number of frames with an accuracy of more than 50\% by using the adjusted-BIC, which adjusts the penalty term of the BIC.

\begin{table}[!t]
    \small
    \centering
    \begin{tabular}{c}
        \begin{minipage}[t]{0.45\hsize}
        \centering
            \begin{tabular}{@{\ }c@{\ \ }|c@{\ \ \ }c@{\ \ \ }c@{\ \ \ }c@{\ \ \  }c@{\ \ }} \bhline
              &  1 &  2 &  3 &  4 & 5+ \\ \hline
            1 &  1 & 16 & 32 & 19 & 52 \\
            2 &  0 & 11 & 14 & 19 & 52 \\
            3 &  0 &  1 &  1 &  5 & 13 \\
            4 &  0 &  0 &  0 &  1 &  3 \\ \bhline
            \end{tabular}

            BIC for FrameNet
        \end{minipage}
        \begin{minipage}[t]{0.45\hsize}
        \centering
            \begin{tabular}{@{\ }c@{\ }|c@{\ \ \ }c@{\ \ \ }c@{\ \ \ }c@{\ \ \  }c@{\ \ }} \bhline
                & 1 &  2 &  3 &  4 &  5+ \\ \hline
            1   & 0 &  0 &  1 & 10 & 109 \\
            2   & 0 &  1 &  1 &  2 &  81 \\
            3   & 0 &  0 &  0 &  1 &  11 \\
            4+  & 0 &  0 &  0 &  0 &  23 \\ \bhline
            \end{tabular}

            BIC for PropBank
        \end{minipage} \\ \\

        \begin{minipage}[t]{0.45\hsize}
        \centering
            \begin{tabular}{@{\ }c@{\ \ }|c@{\ \ \ }c@{\ \ \ }c@{\ \ \ }c@{\ \ \ }c@{\ \ }} \bhline
              &  1 &  2 & 3 & 4 & 5+\\ \hline
            1 & 88 & 17 & 7 & 4 & 4 \\
            2 & 50 & 35 & 5 & 3 & 3 \\
            3 & 10 & 10 & 0 & 0 & 0 \\
            4 &  2 &  1 & 0 & 1 & 0 \\ \bhline
            \end{tabular}

            a-BIC for FrameNet
        \end{minipage}
        \begin{minipage}[t]{0.45\hsize}
        \centering
            \begin{tabular}{@{\ }c@{\ }|c@{\ \ \ }c@{\ \ \ }c@{\ \ \ }c@{\ \ \ }c@{\ \ }} \bhline
                     &  1 &  2 &  3 &  4 & 5+ \\ \hline
            1        & 80 & 35 &  3 &  0 & 2 \\
            2        & 15 & 57 & 12 &  0 & 1 \\
            3        &  1 &  2 &  7 &  1 & 1 \\
            4+       &  0 &  6 &  7 &  3 & 7 \\ \bhline
            \end{tabular}

            a-BIC for PropBank
        \end{minipage}
    \end{tabular}
    \caption{Confusion matrices using BIC and a-BIC for FrameNet and PropBank. Vertical axis represents the number of frames in our dataset, and horizontal axis represents the estimated number of clusters.}
    \label{TB:CM}
\end{table}

In this paper, we focused on the difference of frames that each verb evokes.
That is, we analyzed each verb separately.
However, in FrameNet, frames are shared by several verbs.
For example, the verbs ``support,'' ``prove,'' and ``demonstrate'' can evoke the same \textsc{Evidence} frame. 
To induce FrameNet-style frames, we need to investigate to what extent contextualized word representations capture frames over verbs.

Semantic frame induction requires not only distinguishing the difference of frames that the same verb evokes but also grouping its arguments by the semantic role.
For example, if a sentence contains a verb that evokes the \textsc{Evidence} frame, the sentence contains what is claimed and what supports the claim as its argument.
Contextualized word representations of the arguments will also be useful for grouping arguments by the same roles. 

Furthermore, we only considered verbs as frame-evoking words, but we need to examine whether we can obtain similar results for words with other parts of speech that evoke frames such as nouns. 
These investigations are expected to bring us closer to our goal of automatically constructing high-quality semantic-frame resources.
They can also induce semantic frames for under-resourced languages or specific domains since contextualized word representations do not require human-annotated texts.

\section*{Acknowledgements}
This work was supported by JSPS KAKENHI Grant Numbers 18H03286 and 21K12012.

\bibliographystyle{acl_natbib}
\bibliography{acl2021}

\begin{thebibliography}{32}
\expandafter\ifx\csname natexlab\endcsname\relax\def\natexlab#1{#1}\fi

\bibitem[{Aldous(1985)}]{aldous1985}
David~J Aldous. 1985.
\newblock \href {https://link.springer.com/chapter/10.1007/BFb0099421}
  {Exchangeability and related topics}.
\newblock In \emph{{\'E}cole d'{\'E}t{\'e} de Probabilit{\'e}s de Saint-Flour
  XIII—1983}, pages 1--198.

\bibitem[{Amrami and Goldberg(2018)}]{amrami2018}
Asaf Amrami and Yoav Goldberg. 2018.
\newblock \href {https://www.aclweb.org/anthology/D18-1523/} {Word sense
  induction with neural bi{LM} and symmetric patterns}.
\newblock In \emph{Proceedings of the 2018 Conference on Empirical Methods in
  Natural Language Processing (EMNLP 2018)}, pages 4860--4867.

\bibitem[{Anwar et~al.(2019)Anwar, Ustalov, Arefyev, Ponzetto, Biemann, and
  Panchenko}]{anwar2019}
Saba Anwar, Dmitry Ustalov, Nikolay Arefyev, Simone~Paolo Ponzetto, Chris
  Biemann, and Alexander Panchenko. 2019.
\newblock \href {https://www.aclweb.org/anthology/S19-2018/} {{HHMM} at
  {S}em{E}val-2019 task 2: Unsupervised frame induction using contextualized
  word embeddings}.
\newblock In \emph{Proceedings of the 13th International Workshop on Semantic
  Evaluation (SemEval 2019)}, pages 125--129.

\bibitem[{Arefyev et~al.(2019)Arefyev, Sheludko, Davletov, Kharchev,
  Nevidomsky, and Panchenko}]{arefyev2019}
Nikolay Arefyev, Boris Sheludko, Adis Davletov, Dmitry Kharchev, Alex
  Nevidomsky, and Alexander Panchenko. 2019.
\newblock \href {https://www.aclweb.org/anthology/S19-2004/} {Neural {GRANN}y
  at {S}em{E}val-2019 task 2: A combined approach for better modeling of
  semantic relationships in semantic frame induction}.
\newblock In \emph{Proceedings of the 13th International Workshop on Semantic
  Evaluation (SemEval 2019)}, pages 31--38.

\bibitem[{Baker et~al.(1998)Baker, Fillmore, and Lowe}]{baker1998}
Collin~F Baker, Charles~J Fillmore, and John~B Lowe. 1998.
\newblock \href {https://www.aclweb.org/anthology/P98-1013/} {The {B}erkeley
  {F}rame{N}et project}.
\newblock In \emph{Proceedings of the 36th Annual Meeting of the Association
  for Computational Linguistics and 17th International Conference on
  Computational Linguistics (ACL-COLING 1998)}, pages 86--90.

\bibitem[{Biemann(2006)}]{biemann2006}
Chris Biemann. 2006.
\newblock \href {https://www.aclweb.org/anthology/W06-3812/} {Chinese
  whispers-an efficient graph clustering algorithm and its application to
  natural language processing problems}.
\newblock In \emph{Proceedings of TextGraphs: the First Workshop on Graph Based
  Methods for Natural Language Processing}, pages 73--80.

\bibitem[{Blei et~al.(2003)Blei, Ng, and Jordan}]{blei2003}
David~M Blei, Andrew~Y Ng, and Michael~I Jordan. 2003.
\newblock \href
  {https://www.jmlr.org/papers/volume3/blei03a/blei03a.pdf?TB_iframe=true&width=370.8&height=658.8}
  {Latent dirichlet allocation}.
\newblock \emph{Journal of Machine Learning Research}, 3(Jan):993--1022.

\bibitem[{Cheung and Penn(2013)}]{cheung2013}
Jackie Chi~Kit Cheung and Gerald Penn. 2013.
\newblock \href {https://www.aclweb.org/anthology/P13-1121/} {Towards robust
  abstractive multi-document summarization: A caseframe analysis of centrality
  and domain}.
\newblock In \emph{Proceedings of the 51st Annual Meeting of the Association
  for Computational Linguistics (ACL 2013)}, pages 1233--1242.

\bibitem[{Devlin et~al.(2019)Devlin, Chang, Lee, and Toutanova}]{devlin2019}
Jacob Devlin, Ming-Wei Chang, Kenton Lee, and Kristina Toutanova. 2019.
\newblock \href {https://www.aclweb.org/anthology/N19-1423/} {{BERT}:
  Pre-training of deep bidirectional transformers for language understanding}.
\newblock In \emph{Proceedings of the 2019 Conference of the North {A}merican
  Chapter of the Association for Computational Linguistics: Human Language
  Technologies (NAACL-HLT 2019)}, pages 4171--4186.

\bibitem[{Fillmore(2006)}]{fillmore2006}
Charles~J Fillmore. 2006.
\newblock \href
  {https://d1wqtxts1xzle7.cloudfront.net/56172451/cognitive-linguistics-basics-readings-dirk-geeraerts.pdf?1522158117=&response-content-disposition=inline\%3B+filename\%3DCognitive_linguistics_basics_readings_di.pdf&Expires=1621521870&Signature=TB2rOARBsRMQfgQsU59qLv4UVckZTR-xeKtTVPOndUBjKZFKxJ5SbfhX3wTEwqQrMltszHOY4~ixjcvJMdoyrlFOk6iUwlBiOBngJfpql~o0NIuW0N7DvPoqHD~viV8QIiiuG-m5nItPoDGZ98j3G7GIC411vS5TxtCHmWqJ2brHaswGruTc9y8w~3v2aLMqd8uwEdd6AsAtlP2HZ0hNRFV1hlW9t4b9d1PBT0tx~tbY3UcjPi~nSx2OXP42Yw0BcJcRZO5x4hrw5nQ5aIN1-CwgiCdbjsecTgpgUcPGjTZjbvRt6A~efizCUHWkQS6JPCZ2IwfNUvEuBaLCvrLlOw__&Key-Pair-Id=APKAJLOHF5GGSLRBV4ZA#page=382}
  {Frame semantics}.
\newblock \emph{Cognitive Linguistics: Basic Readings}, 34:373--400.

\bibitem[{Gangemi et~al.(2016)Gangemi, Recupero, Mongiov{\`\i}, Nuzzolese, and
  Presutti}]{gangemi2016}
Aldo Gangemi, Diego~Reforgiato Recupero, Misael Mongiov{\`\i}, Andrea~Giovanni
  Nuzzolese, and Valentina Presutti. 2016.
\newblock \href {https://daneshyari.com/article/preview/571785.pdf}
  {Identifying motifs for evaluating open knowledge extraction on the web}.
\newblock \emph{Knowledge-Based Systems}, 108:33--41.

\bibitem[{Hadiwinoto et~al.(2019)Hadiwinoto, Ng, and Gan}]{hadiwinoto2019}
Christian Hadiwinoto, Hwee~Tou Ng, and Wee~Chung Gan. 2019.
\newblock \href {https://www.aclweb.org/anthology/D19-1533/} {Improved word
  sense disambiguation using pre-trained contextualized word representations}.
\newblock In \emph{Proceedings of the 2019 Conference on Empirical Methods in
  Natural Language Processing and the 9th International Joint Conference on
  Natural Language Processing (EMNLP-IJCNLP 2019)}, pages 5297--5306.

\bibitem[{Kawahara et~al.(2014)Kawahara, Peterson, Popescu, and
  Palmer}]{kawahara2014}
Daisuke Kawahara, Daniel Peterson, Octavian Popescu, and Martha Palmer. 2014.
\newblock \href {https://www.aclweb.org/anthology/E14-1007/} {Inducing
  example-based semantic frames from a massive amount of verb uses}.
\newblock In \emph{Proceedings of the 14th Conference of the {E}uropean Chapter
  of the Association for Computational Linguistics (EACL 2014)}, pages 58--67.

\bibitem[{Lan et~al.(2020)Lan, Chen, Goodman, Gimpel, Sharma, and
  Soricut}]{lan2019}
Zhenzhong Lan, Mingda Chen, Sebastian Goodman, Kevin Gimpel, Piyush Sharma, and
  Radu Soricut. 2020.
\newblock \href {https://openreview.net/pdf?id=H1eA7AEtvS} {{ALBERT}: A lite
  {BERT} for self-supervised learning of language representations}.
\newblock In \emph{Proceedings of the International Conference on Learning
  Representations (ICLR 2020)}.

\bibitem[{Liu(2019)}]{liu2019a}
Yang Liu. 2019.
\newblock \href {https://arxiv.org/pdf/1903.10318.pdf} {Fine-tune {BERT} for
  extractive summarization}.
\newblock {\it ar{X}iv preprint ar{X}iv:1903.10318}.

\bibitem[{Liu et~al.(2019)Liu, Ott, Goyal, Du, Joshi, Chen, Levy, Lewis,
  Zettlemoyer, and Stoyanov}]{liu2019b}
Yinhan Liu, Myle Ott, Naman Goyal, Jingfei Du, Mandar Joshi, Danqi Chen, Omer
  Levy, Mike Lewis, Luke Zettlemoyer, and Veselin Stoyanov. 2019.
\newblock \href {https://openreview.net/pdf?id=SyxS0T4tvS} {Ro{BERT}a: A
  robustly optimized {BERT} pretraining approach}.
\newblock {\it ar{X}iv preprint ar{X}iv:1907.11692}.

\bibitem[{Maaten and Hinton(2008)}]{maaten2008}
Laurens van~der Maaten and Geoffrey Hinton. 2008.
\newblock \href
  {https://www.jmlr.org/papers/volume9/vandermaaten08a/vandermaaten08a.pdf}
  {Visualizing data using t-sne}.
\newblock \emph{Journal of Machine Learning Research}, 9:2579--2605.

\bibitem[{Materna(2012)}]{materna2012}
Ji{\v{r}}{\'i} Materna. 2012.
\newblock \href
  {https://nlp.fi.muni.cz/projekty/lda-frames/static/ldaf/lda-frames.pdf}
  {Lda-frames: An unsupervised approach to generating semantic frames}.
\newblock In \emph{International Conference on Intelligent Text Processing and
  Computational Linguistics (CICLing 2012)}, pages 376--387.

\bibitem[{Mikolov et~al.(2013)Mikolov, Sutskever, Chen, Corrado, and
  Dean}]{mikolov2013}
Tomas Mikolov, Ilya Sutskever, Kai Chen, Greg~S Corrado, and Jeff Dean. 2013.
\newblock \href
  {https://papers.nips.cc/paper/2013/file/9aa42b31882ec039965f3c4923ce901b-Paper.pdf}
  {Distributed representations of words and phrases and their
  compositionality}.
\newblock In \emph{Advances in Neural Information Processing Systems (NIPS
  2013)}, pages 3111--3119.

\bibitem[{Palmer et~al.(2005)Palmer, Gildea, and Kingsbury}]{palmer2005}
Martha Palmer, Daniel Gildea, and Paul Kingsbury. 2005.
\newblock \href {https://www.aclweb.org/anthology/J05-1004/} {The proposition
  bank: An annotated corpus of semantic roles}.
\newblock \emph{Computational linguistics}, 31(1):71--106.

\bibitem[{Peters et~al.(2018)Peters, Neumann, Iyyer, Gardner, Clark, Lee, and
  Zettlemoyer}]{peters2018}
Matthew Peters, Mark Neumann, Mohit Iyyer, Matt Gardner, Christopher Clark,
  Kenton Lee, and Luke Zettlemoyer. 2018.
\newblock \href {https://www.aclweb.org/anthology/N18-1202/} {Deep
  contextualized word representations}.
\newblock In \emph{Proceedings of the 2018 Conference of the North {A}merican
  Chapter of the Association for Computational Linguistics: Human Language
  Technologies (NAACL-HLT 2018)}, pages 2227--2237.

\bibitem[{QasemiZadeh et~al.(2019)QasemiZadeh, Petruck, Stodden, Kallmeyer, and
  Candito}]{qasemizadeh2019}
Behrang QasemiZadeh, Miriam R.~L. Petruck, Regina Stodden, Laura Kallmeyer, and
  Marie Candito. 2019.
\newblock \href {https://www.aclweb.org/anthology/S19-2003/} {{S}em{E}val-2019
  task 2: Unsupervised lexical frame induction}.
\newblock In \emph{Proceedings of the 13th International Workshop on Semantic
  Evaluation (SemEval 2019)}, pages 16--30.

\bibitem[{Radford et~al.(2019)Radford, Wu, Child, Luan, Amodei, and
  Sutskever}]{radford2019}
Alec Radford, Jeffrey Wu, Rewon Child, David Luan, Dario Amodei, and Ilya
  Sutskever. 2019.
\newblock \href
  {https://d4mucfpksywv.cloudfront.net/better-language-models/language_models_are_unsupervised_multitask_learners.pdf}
  {Language models are unsupervised multitask learners}.
\newblock \emph{OpenAI Blog}, 1(8).

\bibitem[{Ribeiro et~al.(2019)Ribeiro, Mendon{\c{c}}a, Ribeiro, Martins~de
  Matos, Sardinha, Santos, and Coheur}]{ribeiro2019}
Eug{\'e}nio Ribeiro, V{\^a}nia Mendon{\c{c}}a, Ricardo Ribeiro, David
  Martins~de Matos, Alberto Sardinha, Ana~L{\'u}cia Santos, and Lu{\'\i}sa
  Coheur. 2019.
\newblock \href {https://www.aclweb.org/anthology/S19-2019/}
  {{L}2{F}/{INESC}-{ID} at {S}em{E}val-2019 task 2: Unsupervised lexical
  semantic frame induction using contextualized word representations}.
\newblock In \emph{Proceedings of the 13th International Workshop on Semantic
  Evaluation (SemEval 2019)}, pages 130--136.

\bibitem[{Schwarz(1978)}]{schwarz1978}
Gideon Schwarz. 1978.
\newblock \href
  {https://projecteuclid.org/journals/annals-of-statistics/volume-6/issue-2/Estimating-the-Dimension-of-a-Model/10.1214/aos/1176344136.full}
  {Estimating the dimension of a model}.
\newblock \emph{The Annals of Statistics}, 6(2):461--464.

\bibitem[{Shen and Lapata(2007)}]{shen2007}
Dan Shen and Mirella Lapata. 2007.
\newblock \href {https://www.aclweb.org/anthology/D07-1002/} {Using semantic
  roles to improve question answering}.
\newblock In \emph{Proceedings of the 2007 joint conference on empirical
  methods in natural language processing and computational natural language
  learning (EMNLP-CoNLL 2007)}, pages 12--21.

\bibitem[{Ustalov et~al.(2018)Ustalov, Panchenko, Kutuzov, Biemann, and
  Ponzetto}]{ustalov2018}
Dmitry Ustalov, Alexander Panchenko, Andrey Kutuzov, Chris Biemann, and
  Simone~Paolo Ponzetto. 2018.
\newblock \href {https://www.aclweb.org/anthology/P18-2010/} {Unsupervised
  semantic frame induction using triclustering}.
\newblock In \emph{Proceedings of the 56th Annual Meeting of the Association
  for Computational Linguistics (ACL 2018)}, pages 55--62.

\bibitem[{Vaswani et~al.(2017)Vaswani, Shazeer, Parmar, Uszkoreit, Jones,
  Gomez, Kaiser, and Polosukhin}]{vaswani2017}
Ashish Vaswani, Noam Shazeer, Niki Parmar, Jakob Uszkoreit, Llion Jones,
  Aidan~N Gomez, {\L}ukasz Kaiser, and Illia Polosukhin. 2017.
\newblock \href
  {https://papers.nips.cc/paper/2017/file/3f5ee243547dee91fbd053c1c4a845aa-Paper.pdf}
  {Attention is all you need}.
\newblock In \emph{Advances in Neural Information Processing Systems (NIPS
  2017)}, pages 5998--6008.

\bibitem[{Yang and Mitchell(2017)}]{yang2017}
Bishan Yang and Tom Mitchell. 2017.
\newblock \href {https://www.aclweb.org/anthology/D17-1128/} {A joint
  sequential and relational model for frame-semantic parsing}.
\newblock In \emph{Proceedings of the 2017 Conference on Empirical Methods in
  Natural Language Processing (EMNLP 2017)}, pages 1247--1256.

\bibitem[{Yang et~al.(2019)Yang, Dai, Yang, Carbonell, Salakhutdinov, and
  Le}]{yang2019}
Zhilin Yang, Zihang Dai, Yiming Yang, Jaime Carbonell, Russ~R Salakhutdinov,
  and Quoc~V Le. 2019.
\newblock \href
  {https://papers.nips.cc/paper/2019/file/dc6a7e655d7e5840e66733e9ee67cc69-Paper.pdf}
  {{XLN}et: Generalized autoregressive pretraining for language understanding}.
\newblock In \emph{Advances in Neural Information Processing Systems (NeurlIPS
  2019)}, pages 5754--5764.

\bibitem[{Yong and Timponi~Torrent(2020)}]{yong2020}
Zheng~Xin Yong and Tiago Timponi~Torrent. 2020.
\newblock \href {https://www.aclweb.org/anthology/2020.lrec-1.431/}
  {Semi-supervised deep embedded clustering with anomaly detection for semantic
  frame induction}.
\newblock In \emph{Proceedings of The 12th Language Resources and Evaluation
  Conference (LREC 2020)}, pages 3509--3519.

\bibitem[{Zhu et~al.(2019)Zhu, Xia, Wu, He, Qin, Zhou, Li, and Liu}]{zhu2019}
Jinhua Zhu, Yingce Xia, Lijun Wu, Di~He, Tao Qin, Wengang Zhou, Houqiang Li,
  and Tieyan Liu. 2019.
\newblock \href {https://openreview.net/forum?id=Hyl7ygStwB} {Incorporating
  {BERT} into neural machine translation}.
\newblock In \emph{Proceedings of the International Conference on Learning
  Representations (ICLR 2019)}.

\end{thebibliography}


\end{document}